\documentclass[10pt,twocolumn]{article} 
\usepackage{simpleConference}
\usepackage{times}
\usepackage{graphicx}
\usepackage{amssymb}
\usepackage{url,hyperref}
\usepackage{graphicx}
\usepackage{caption}
\usepackage{makecell}
\usepackage[normalem]{ulem}

\usepackage{booktabs}
\usepackage{mathtools}
\usepackage{float}
\usepackage{array}
\usepackage{multirow}
\usepackage{tabularx}
\usepackage{colortbl}
\usepackage{hhline}
\usepackage[table]{xcolor}
\usepackage{array,hhline}
\RequirePackage[table]{xcolor}
\usepackage{array,hhline}
\usepackage{marvosym}
\usepackage{pgf}
\usepackage{indentfirst}
\usepackage{threeparttable}
\usepackage{amsmath}

\hypersetup{
    colorlinks=true,
    linkcolor=blue,
    filecolor=magenta,      
    urlcolor=cyan,
}
\usepackage{authblk}

\begin{document}
\title{HGR-Net: A Fusion Network for Hand Gesture Segmentation and Recognition}
\author{{Amirhossein Dadashzadeh$^{1}$}, {Alireza Tavakoli Targhi$^{2}$}\thanks{a\_tavakoli@sbu.ac.ir} , {Maryam Tahmasbi$^{2}$}, {Majid Mirmehdi$^{1}$}}

\affil{ Department of Computer Science, University of Bristol, Bristol, UK$^{1}$ \\
{Department of Computer Science, Shahid Beheshti University, Tehran, Iran}$^{2}$}


\maketitle
\thispagestyle{empty}

\begin{abstract}
We propose a two-stage convolutional neural network (CNN) architecture for robust recognition of hand gestures,
called HGR-Net, where the first stage performs accurate semantic segmentation to determine hand regions, and the second
stage identifies the gesture. The segmentation stage architecture is based on the combination of fully convolutional residual
network and atrous spatial pyramid pooling. Although the segmentation sub-network is trained without depth information, it is
particularly robust against challenges such as illumination variations and complex backgrounds. The recognition stage deploys
a two-stream CNN, which fuses the information from the red-green-blue and segmented images by combining their deep
representations in a fully connected layer before classification. Extensive experiments on public datasets show that our
architecture achieves almost as good as state-of-the-art performance in segmentation and recognition of static hand gestures,
at a fraction of training time, run time, and model size. Our method can operate at an average of 23 ms per frame.

\end{abstract}

\maketitle

\section{Introduction}\label{sec1}

\label{intro}
Gestures are an integral part of social interaction which can help us to convey information and express our thoughts and feelings more effectively.  
Hand gestures enable an intuitive and natural means of connection between humans, and can also serve well for interaction between humans and machines.
Indeed, recognizing hand gestures has long been an active area of research in visual pattern analysis with a wide range of applications, including human-computer interaction \cite{Rautaray2015}, sign language communication \cite{Starner1998,Cooper2011}, virtual reality \cite{Xu2006}. 


{Hand Gesture Recognition (HGR) has been tackled by wearable sensors \cite{Dipietro2008} and vision sensors \cite{pisharady2015recent,Rautaray2015} to identify static and/or dynamic gestures. A vision-based approach is preferable as it offers a remote, contact-free and still affordable HGR system. The traditional methods from the pre-deep learning era used hand-crafted shape and appearance features for static gestures \cite{plouffe2016static}, and added motion features to recognize dynamic gestures in video sequences \cite{de2016skeleton,lu2016dynamic, singha2018dynamic}}. Recently, deep learning models have achieved great success in a wide range of computer vision tasks, including image classification \cite{Krizhevsky2012}, object detection \cite{Ren2015} and semantic segmentation \cite{Long2015,Guo2018}. These successes have motivated many researchers to develop HGR systems based on deep learning algorithms \cite{chevtchenko2018convolutional,Molchanov2015,Oyedotun2017}. 

An important preprocessing step in HGR is the segmentation of hand regions - a highly challenging task, due to the variability of the image background, ethnicity color differences, shadows, and other illumination variations.
One potential solution to deal with such difficulties is 
to use depth sensors, such as the Microsoft Kinect \cite{Dal2012}.
However, depth-sensing devices might not be suitable for all environments, especially in outdoor scenarios, and add to the computational burden.

\begin{figure*}[h] 
\centering
\includegraphics[scale=0.14]{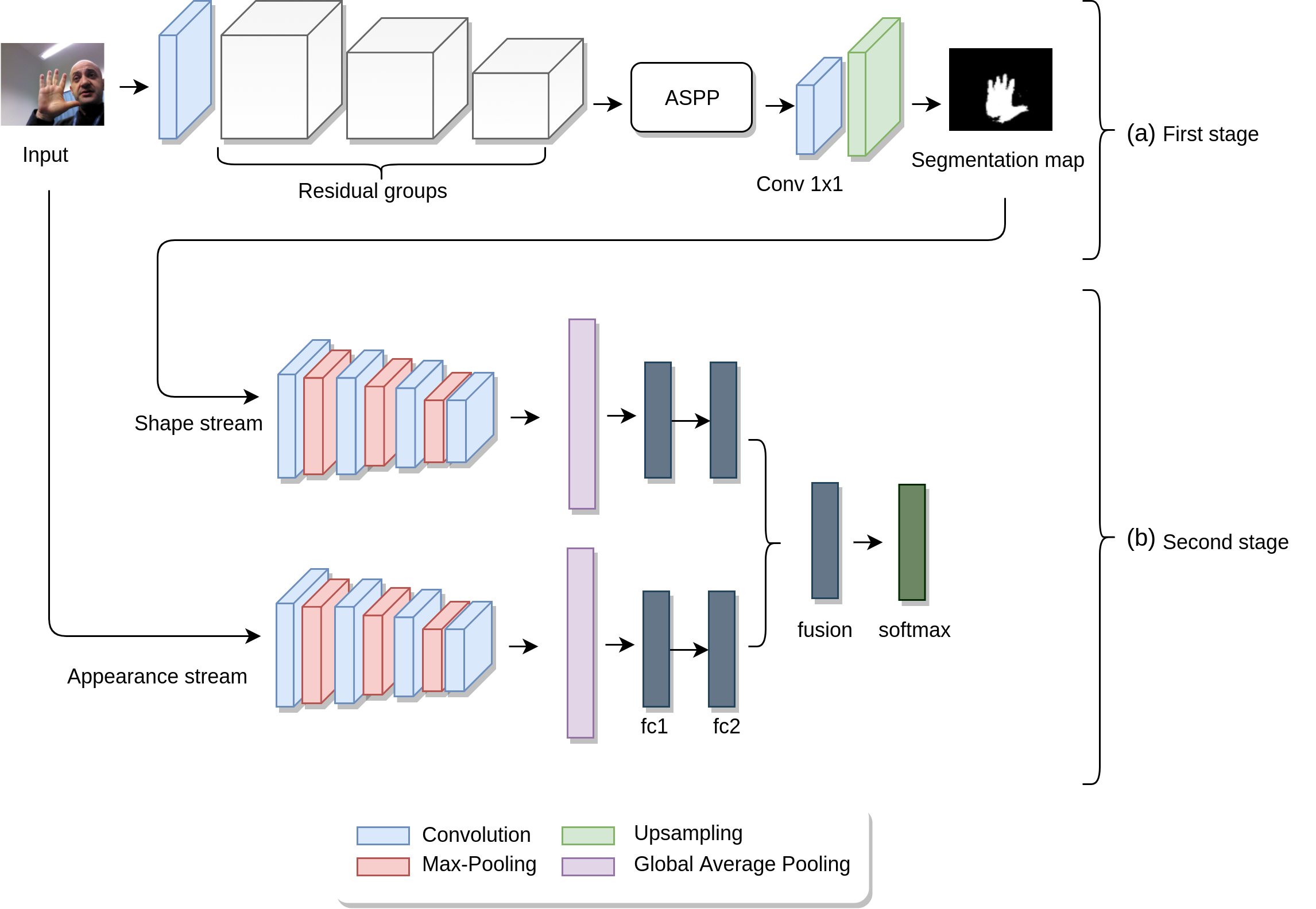}
\caption{\scriptsize\textit{{The proposed HGR-Net: (a) a hand segmentation network is first trained to segment hand regions from the background, (b) a shape stream and an appearance stream are then trained separately to capture corresponding information, which are then combined in a fusion layer before a softmax classifier.} }}
 
\label{fig:HGR-Net}
\end{figure*}

In this paper, we propose a novel HGR system using a deep learning framework, where we design a two-stage convolutional neural network (CNN) architecture for segmentation and recognition of hand gestures. An overview of the network is shown
in Fig. \ref{fig:HGR-Net}.

At the first stage of the network, a deep architecture based on {fully convolutional residual network} is used to segment the hand out of the image. This stage adopts an Atrous  Spatial  Pyramid Pooling (ASPP) module \cite{Chen2017}, which encodes global contextual information at multiple scales.
In the second stage, we build a two-stream CNN 
 to learn feature representations for {the input RGB images and their corresponding hand segmentation maps from the first stage} separately, which are then combined with a late fusion approach. This architecture bears similarity to other recent multi-stream CNNs \cite{eitel2015multimodal,feichtenhofer2016convolutional,zhu2016discriminative}. However, in this work we consider the relationship between the RGB images and {the segmentation map}, instead of {requiring multimodal} RGB and depth or motion (optical flow). The main advantage of this two-stage architecture is that it exploits the RGB image and {the segmentation map} at training time, 
{and recognizes the hand directly from only the 2D RGB frames at testing time, with an average processing rate of 23ms per frame.}

To improve the generalization capability of HGR-Net significantly, we present an efficient data augmentation strategy by employing both online and offline data augmentation. {The proposed method is evaluated on two benchmark datasets, i.e.  OUHANDS \cite{Matilainen2016} and HGR1 \cite{HGR1}.}
Our experiments show that by using a robust hand segmentation/recognition architecture, supported by an efficient data augmentation technique, we can develop a robust {RGB-based} HGR system which has excellent performance in uncontrolled and unseen scenarios, at much lower training and computational costs than the current state of the art\footnote{The code and trained models of our paper will be made available on github.}.


Next, in Section \ref{Sec:Related-Works}, we describe related works. In Section \ref{Sec:method},  our proposed methodology, encapsulating the design of HGR-Net, is outlined. The experimental results are discussed in Section \ref{Sec:Experimental}, and finally we conclude the paper in Section \ref{Sec:Conclusion}.

\section{Related Works}\label{Sec:Related-Works}

{\bf Hand Segmentation - } Segmentation of hands plays an important role in many computer vision applications, such as  hand tracking \cite{Qian2014}, hand pose estimation \cite{tompson2014real} and gesture recognition \cite{Rautaray2015}. In recent years, skin segmentation has been widely used by several approaches for hand segmentation \cite{Han2006,Nalepa2014,Joshi2015}. Among {such works} 
color-based methods have been popular with {many applying} threshold-based techniques on different color spaces to achieve the best skin segmentation performance {\cite{Argyros2004,kawulok2014skin,sawicki2015human}}. 

Machine learning techniques for skin segmentation, such as  \cite{Han2006,Bhoyar2010,khan2012} utilize a classifier trained on 
skin (hand) and non-skin (background) classes.
{In \cite{Kawulok2013,Kawulok2014}, Kawulok and co-workers} extracted textural features from skin probability maps and then propagated them in a spatial analysis scheme. 
Hettiarachchi {and Peters} \cite{Hettiarachchi2016} 
{built {their} skin classifier training data set with skin
and skin-like classes instead of skin and non-skin classes in order to obtain a more balanced class distribution in the training dataset. They segmented the skin candidate regions using a Voronoi-based image segmentation technique and then {classified} them using a multi-manifold-based skin classifier.}

{These previous approaches have significant drawbacks in that they} 
{do not perform well} when the background color is similar to skin color, and {that} they are seriously affected by lighting conditions.

To overcome {such} limitations,
many {works have taken} advantage of depth information{, such as} \cite{Shotton2011,palacios2013human,Qian2014,Kang2017}. For example, Kang \textit{et al.} \cite{Kang2017} proposed a two-stage random decision forest (RDF) classifier, {where the first stage detects hands by processing the RDF on an entire depth map, and the second stage segments hands by applying the RDF in the detected {regions}.}

{\bf {Hand Gesture Recognition} - } Numerous methods for recognizing hand gestures in static images have been proposed over the years, e.g
\cite{Pisharady2013,plouffe2016static,chevtchenko2018convolutional}.
Among {traditional} feature-based methods, 
Pisharady \textit{et al}. \cite{Pisharady2013} utilized a Bayesian model of visual attention to identify hand regions within images. {Hand gestures were} recognized using {low level (color) and high level (shape, texture)}  features of the hand areas, with a Support Vector Machines (SVM) classifier.
Priyal and Bora \cite{priyal2013robust} used geometry based normalizations and Krawtchouk moment features to represent binary silhouettes of hands in a rotation invariant manner.
 Nalepa \textit{et al}. \cite{Nalepa2014} proposed a parallel algorithm for hand shape classification which performed in real-time over the segmented hand gestures. {They} combined shape contexts with appearance-based techniques to increase the classification score.
 A combination of graph-based characteristics and {shape-based} features, such as Canny edges,  {was} proposed by Avraam \cite{avraam2014static} for modeling static hand gestures.

Depth cameras have also been applied to the HGR problem, e.g. \cite{palacios2013human,Marin2014,memo2018head,plouffe2016static}. In \cite{Marin2014}, the authors {combined}  depth features and skeletal data descriptors using Leap Motion and Kinect devices. An SVM classifier with {a} {Gaussian} radial basis function kernel was then used to recognize the performed gestures.
In \cite{plouffe2016static}, the k-curvature algorithm {was} employed to describe hand poses through extracted contours. Then, a dynamic time warping algorithm {was} used for recognition, where the {gesture under analysis was} compared with a series of pre-recorded reference gestures. This technique {was} used for the classification of static and dynamic hand gestures.

{\bf Deep learning-based methods - }  
Barros \textit{et al}. \cite{barros2014multichannel}
developed {a convolutional neural network {for hand pose analysis} with three channels, where one takes a gray-scale image and the second and third receive the {edge maps resulting from the application of the} Sobel filter in horizontal and vertical directions, respectively.}

 Liang \textit{et al}. \cite{Liang2016} proposed a multi-view framework for recognizing hand gestures using point clouds captured by a depth sensor. They used CNNs as feature extractors and finally employed an SVM classifier to classify hand gestures.
 Oyedotun and Khashman \cite{Oyedotun2017} utilized a CNN and stacked denoising autoencoder for recognizing 24 American Sign Language hand gestures. They evaluated their recognition performance on a public database, which was collected under controlled background and lighting conditions, {reaching recognition rates of 91.33\% and 92.83\% by models called CNN1 and SDAE3, respectively.} 
 Recently, Chevtchenko \textit{et al}. \cite{chevtchenko2018convolutional} presented a new architecture based on the combination of a CNN and traditional feature extractors including Hu moments, Gabor filters and Zernike moments.
 The authors evaluated their method on binary, depth and grayscale images, using data where the background was either already removed or ignored by operating on the hand as the closest object to the camera.
  Moreover, their experimental results {on the OUHANDS dataset  \cite{Matilainen2016}} cannot really represent the generalization capacity of their model, since they apply a leave-one-out cross validation approach {on the entire dataset that contains training, validation, and test sets.} 
  Finally, their method cannot be directly compared against as they do not report their results on the standard test set of the dataset.

{{In conclusion,} there is still a lack of a highly accurate solution to the problem of non-depth based, static, hand gesture recognition in uncontrolled environments. In this article, we address this issue by proposing a two-stage 
CNN architecture, which enables the full exploitation of the shape and appearance information from single RGB images.}

\section{Proposed {Method}}\label{Sec:method} 

The structure of the proposed HGR-Net is shown in Fig. \ref{fig:HGR-Net}. In summary, the network consists of {three CNNs, operating over two stages}.
In the first stage,
{the candidate hand regions are segmented in the RGB image and}
used as input to the second stage (Section \ref{SubSec:HandSeg-Stage}). The second stage is composed of two streams, one for the input RGB image and one for the segmentation map from the first stage (Section \ref{SubSec:HandRec-Stage}). Each of the streams consists of a deep CNN which are converged in a fully-connected layer and a softmax classifier. To train the HGR-Net, we proceed in a stage-wise manner which is common practice in {multi-stream architectures} 
(Section \ref{Subsec:training-Paradigm}).

\subsection{{Hand Segmentation}}
\label{SubSec:HandSeg-Stage}
{Most deep learning techniques that have been successfully used for semantic segmentation} are based on Fully Convolutional Networks (FCN) \cite{Long2015,Chen2017}. The main idea in FCN-based networks is that they remove fully-connected layers to create an efficient end-to-end learning algorithm. In this work, we take advantage of this idea to build a powerful deep network for hand segmentation. {Our network, explained in detail below,} consists of two main parts: (1) a deep fully convolutional residual network for learning useful {representations,} and (2) an
{ASPP module} to encode multi-scale context by adopting multiple dilation rates. 

{\bf Fully convolutional residual network - } Theoretically, increasing network depth should increase the capacity and representation of the network \cite{Goodfellow2016}. However, in practice this is impossible due to many limitations, such as the risk of overfitting, {vanishing gradients}, and difficulty in {efficient} optimization. So increasing network layers to make a deeper network does not work by simply stacking layers together. To overcome these problems, He \textit{et al}. \cite{He2016} proposed a residual learning technique to ease the training of networks {that enables} them to be substantially deeper. This technique allows gradients to flow across multiple layers during the training process by using identity mappings as skip connections. We take advantage of residual learning to {construct} a deep CNN with 28 convolution layers.
 \begin{figure}[h]
\centering
\includegraphics[scale=0.15]{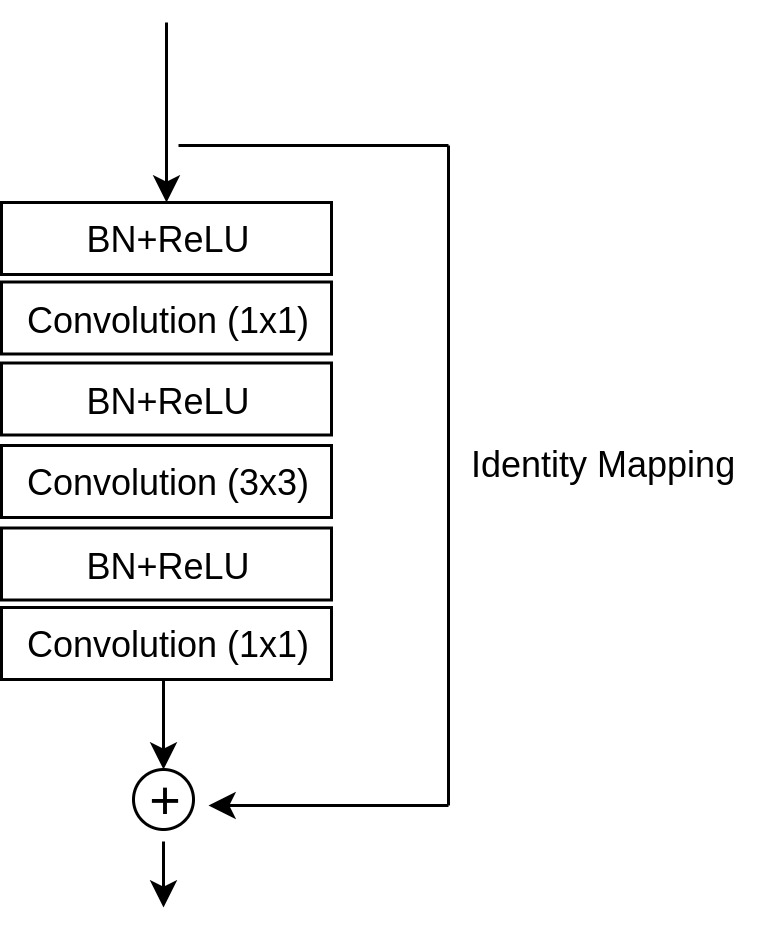}
\caption{\scriptsize\textit{{The structure of residual unit with bottleneck used in the proposed method. BN denotes Batch Normalization.} }}
\label{fig:ResidualUnit}
\end{figure}
As shown in { Fig. \ref{fig:HGR-Net}(a) and Table \ref{table:Residual-Net},} we consider 3 residual groups {where} each group contains 3 residual units, stacked together. The structure of these units is shown in Fig. \ref{fig:ResidualUnit}. Each unit can be defined as
\begin{equation} 
  y_{k}=h(x_{k})+F(x_{k};W_{k}) , 
\end{equation}
{where} $x_{k}$ and $y_{k}$ represent the input and output of the {$k^{th}$} unit, $h(x_{k})$ is an identity mapping and $F$ is the residual function defined by weights $W_{k}$.

As {seen} in Fig. \ref{fig:ResidualUnit}, we use a bottleneck architecture for each residual unit with two $1\times1$  convolution layers. This architecture
reduces the dimensionality and then restores it. The downsampling operation is performed by the first $1\times1$ convolution in the second and third residual groups with a stride of 2.

\begin{table}[!b]
\caption{\scriptsize\textit{Our fully convolutional residual network architecture with 28 convolution layers in the first stage of HGR-Net.\label{table:Residual-Net}}}
\scalebox{0.8}{
{\begin{tabular*}{25pc}{@{\extracolsep{\fill}}lll@{}}\toprule
{\bf Layer name} & {\bf Output shape} & {\bf Layers} \\

\midrule
        input & $320\times320\times3$ \\
       
        convolution  & $320\times320\times16$ \\
        \\
        ResGroup1 & $320\times320\times32$&$\begin{bmatrix}1&1&8\\3&3&8\\1&1&32\end{bmatrix} \times 3 $ \\
        \\
        ResGroup2 & $160\times160\times64$&$\begin{bmatrix}1&1&16\\3&3&16\\1&1&64\end{bmatrix} \times 3 $, stride=2\\
        \\
        ResGroup3 & $80\times80\times128$&$\begin{bmatrix}1&1&32\\3&3&32\\1&1&128\end{bmatrix} \times 3 $, stride=2 \\
\bottomrule
\end{tabular*}}{}}
\end{table}

\textbf{{Atrous Spatial Pyramid Pooling - }}
Contextual information has been shown to be extremely important for semantic segmentation tasks \cite{galleguillos2010context,liu2015semantic}.
In a CNN, the size of the receptive field can roughly determine how much we need to capture context. Although the wider receptive filed allows us to gather more context, Zhou \textit{et al}. \cite{zhou2014object} showed that the actual size of the receptive fields in a CNN is much smaller than the theoretical size, especially {in higher} level layers.
To solve this problem, Yu and Koltun \cite{Yu2015} {proposed} atrous convolution which can exponentially enlarge the receptive field to capture long-range context without losing spatial resolution and increasing the number of parameters. 

\begin{figure}[h]
\centering
\includegraphics[scale=0.15]{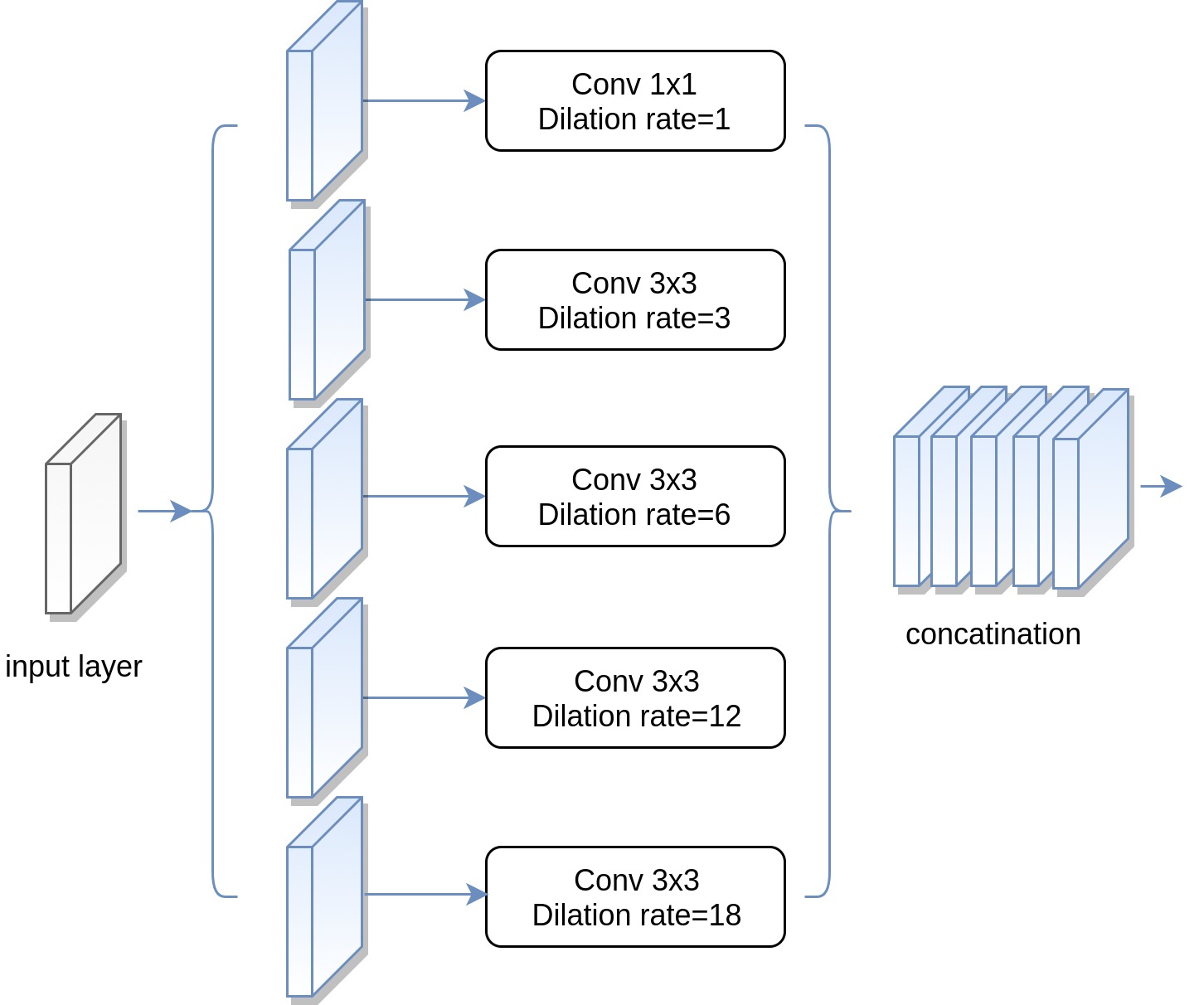}
\caption{\scriptsize\textit{The structure of {our ASPP} module with 5 parallel operations.} }
\label{fig:aspp}
\end{figure}

We combine different scales of contextual information using {an ASPP  module as shown in} Fig. \ref{fig:aspp}. This kind of module has been employed successfully in the state-of-the-art {DeepLabv3 \cite{Chen2017}} 
for semantic segmentation. The ASPP module used in this work 
{has five levels,} a $1\times 1$ convolution and four $3\times 3$ convolutions with atrous rates of 1, 3, 6, 12 and 18 respectively (all with 32 filters). The {resulting} five feature maps are concatenated together. 
Then, the output from the concatenation unit is fed to a $1\times1$ convolution and is upsampled by a standard 4x bilinear interpolation to provide an end-to-end learning algorithm {for hand segmentation (see Fig. \ref{fig:HGR-Net}(a)).}

\subsection{{Hand Gesture Recognition}} \label{SubSec:HandRec-Stage}
{In this stage of our proposed method, we} predict the hand gesture label using a 
{fusion of two CNNs.}
As shown in Fig. \ref{fig:HGR-Net}(b), this stage consists of two separate CNNs with the same architecture where each network exploits the shape-based and appearance-based information respectively, conveyed by {the hand segmentation map from the previous stage} and the RGB image for robust hand gesture recognition. Each  CNN consists of several operations including convolution, pooling and ReLUs, followed by two fully-connected layers {at} the end of the network. The {detailed} structure of these CNNs is shown in Table \ref{table:Rec-CNN}. 

{A fusion function is then deployed to fuse the outputs of the last fully-connected layer of each network} in an element-wise summation manner.
More formally, {let} $f^{S}\in\mathbb{R}^{d}$ and $f^{A}\in\mathbb{R}^{d}$ be the outputs of the last fully connected layers (fc2) from the {first (shape stream) and second (appearance stream)} CNN respectively. The output of the fusion function $f^{sum}$ is 
\begin{equation} 
f^{sum}_{i}= f^{S}_{i} + f^{A}_{i},
\label{Equation:fusion}
\end{equation} 
where $1\leq i\leq d$ and $f^{sum}\in\mathbb{R}^{d}$, {and $d $ is number of the neurons (in our expriments 64 for fc2). The} feature vector $f^{sum}$ is then fed into a softmax classifier for joint supervised learning. It is worth noting that, since our model benefits from a segmentation, it takes {the RGB image as the only input} at testing time.
\begin{table}[!b]
\caption{\scriptsize\textit {The architecture of the CNNs used in the recognition stage (Fig. \ref{fig:HGR-Net}(b). Note that both CNNs have the same architecture, {except that the shape stream CNN uses a one-channel input image.}}\label{table:Rec-CNN}}
{\begin{tabular*}{20pc}{@{\extracolsep{\fill}}lcc@{}}\toprule
{\bf Layer}       &  {\bf Type}   &  {\bf Output Shape} \\ 
\midrule
input        &                         & $320\times320\times3$    \\ 
conv1       & convolution              & $318\times318\times16$   \\
pool1        & max-pooling             & $106\times106\times16$   \\ 
conv2        & convolution             & $104\times104\times32$   \\
pool2        & max-pooling             & $34\times34\times32$     \\
conv3        & convolution             & $32\times32\times64$     \\
pool3        & max-pooling             & $10\times10\times64$     \\
conv4        & convolution             & $8\times8\times128$      \\
pool4        & global average pooling  & $128$          \\
dropout1     & dropout &   $128$          \\
fc1          & fully connected   & $64$          \\
dropout2     & dropout &  $64$          \\

fc2        & fully connected   & $64$          \\ 
\bottomrule
\end{tabular*}}{}
\end{table}

\subsection{{Training Paradigm}} \label{Subsec:training-Paradigm}
\label{Subsec:training-Paradigm}
Let $S = \{( \textbf x_{i}, \textbf m_{i}, \textbf y_{i})$ $\text {for}$ $ i=1,...,N\}$ be the $N$ training samples where $\textbf x_{i} \in \mathbb{R}^{h \times w \times 3}$, $\textbf m_{i} \in \mathbb{R}^{h \times w \times 1}$ and $\textbf y_{i}  \in \mathbb{R}^{c}$ are the RGB images, segmentation masks, and the corresponding image labels in one-hot encoding respectively.
The training paradigm of the HGR-Net is divided {into} three main steps. 

{\it {1) Training the segmentation network:}}
Since {the shape} stream relies on the hand segmentation map, in the first step we train the hand segmentation network (Fig. \ref{fig:HGR-Net}(a)) using RGB images ($\textbf x_i$) and the corresponding hand segmentation masks ($\textbf m_i$). To find the optimal weights of the segmentation network, we need to {minimize,}
\begin{equation}
\min_{ \theta^{seg}} \dfrac{1}{N}\sum_{i=1}^N \mathcal{L}({\mathsf{sigmoid}}(p^{seg}(\textbf x_i,\theta^{seg})),\textbf m_i ),
\end{equation}
where $\theta^{seg}=\{{\text W^{seg},\text b^{seg}} \}$ are the parameters 
(weights and biases) of the segmentation network and $p^{seg}(.)$ is the function that maps the $\textbf x_i$ to the class scores $\textbf s^{seg}$, {where}  $\textbf s^{seg} \in \mathbb{R}^{h \times w \times 1} $.
We use the $\mathsf{sigmoid}$ function to produce the segmentation probability map by squashing the values of $\textbf s^{seg}$ in the range (0, 1).
The loss function $\mathcal{L}(.)$ is defined as 
\begin{equation}
    \mathcal{L}(\textbf p,\textbf y)=-[ \textbf y \log (\textbf p) + (1- \textbf y)\log (1-\textbf p)],
    \label{Eq:: CE-loss}
\end{equation}
where $\textbf y$ is the true class (in this step $\textbf y$=$\textbf m_i$), and $\textbf p$ is the predicted probability for that class (in this step $\textbf p= \mathsf{sigmoid}$($\textbf s^{seg}$)). We use this loss function in all our training steps.

{{\it 2) Training the stream {networks}:} 
After training the segmentation network, we keep its trained weights to train the shape stream using 
RGB images ($\textbf x_i$) and {their corresponding one-hot vectors ($\textbf y_i$)}}. To find the optimal weights of the shape stream network, we need to minimize the loss function $\mathcal{L}$, described in this step as
\begin{equation}
\min_{ \theta^{S}} \dfrac{1}{N}\sum_{i=1}^N \sum_{c=1}^C \mathcal{L}(\mathsf{softmax}(p^{rec}( \text x_i,\theta^{seg},\theta^{S})),\text y_i ),
\end{equation}
where $\theta^{S}=\{{\text W^S,\text b^S} \}$ are the parameters of the shape stream network and $p^{rec}(.)$ is the function that maps the $\textbf x_i$ to the class scores $\textbf s^{rec}$, {where} $\textbf s^{rec} \in \mathbb{R}^{c}$.
By using the softmax classifier, we map  $\textbf s^{rec}$ to a probability distribution over $C$ classes. In this step, the softmax classifier is replaced with the last fully connected layer (fc2). 

We also train the appearance stream in a similar fashion, except that this stream does not rely on the hand segmentation network weights, as shown in Fig. \ref{fig:HGR-Net}(b). {

{{\it 3) Training the fusion network:} }
In the last step, we {initialize the HGR-Net weights} with the weights of the pre-trained CNNs from previous steps and we train the whole network (HGR-Net) by minimizing, 
\begin{equation}
\min_{ \theta^{S},\theta^{A},\theta^{F}} \dfrac{1}{N}\sum_{i=1}^N \sum_{c=1}^C \mathcal{L}(\mathsf{softmax}(p^{rec}( \text x_i,\theta^{seg},\theta^{S},\theta^{A},\theta^{F})),\text y_i ),
\label{eq:: loss-HGR-Net}
\end{equation}
where $\theta^{A}=\{{\text W^A,\text b^A} \}$ and $\theta^{F}=\{{\text W^F,\text b^F} \}$ are the parameters of the appearance stream and fusion network after fusion layer, respectively.

Note that in this step, training can be performed by optimizing only the weights after fc2 layers and the remaining weights can be frozen.}

\section{{Experimental Evaluation}}\label{Sec:Experimental}

{\bf Datasets - } 
{To the best of our knowledge, there are currently} very few publicly available datasets for static hand gesture {analysis} that include hand segmentation groundtruth and were captured under challenging {situations}. For instance, datasets such as Kinect Leap \cite{Marin2014} or Senz3D \cite{memo2018head} contain complex backgrounds and face-hand occlusions, but they lack segmentation groundtruth and do not provide a sufficient number of subjects. {Moreover, these datasets contain depth maps which are not {aligned} with their corresponding RGB images, {making the generation of segmentation annotations from these depth maps unreliable.}}

\noindent We evaluate our proposed method on two benchmark datasets:

{OUHANDS \cite{Matilainen2016} -} This dataset contains 10 different hand gestures from 23 subjects, and is split into training, validation and testing sets with 1200, 400 and 1000 images respectively. All sets come with corresponding segmentation masks. Each set is performed by different subjects and the images were captured under very challenging situations, such as variations in illumination, complex backgrounds and face-hand occlusions with various hand shapes and sizes. {We use this dataset {to evaluate 
HGR-Net for both} hand segmentation and hand gesture recognition.}

{HGR1 \cite{HGR1} -} This dataset contains 899 RGB images of 12 individuals performing 25 different hand gestures, collected under uncontrolled backgrounds without any face-hand occlusion. 
{We use HGR1 for evaluating our hand segmentation accuracy only for two reasons. First, the database has only relatively few samples per class, hence it is not particularly useful to evaluate our recognition model. Second, as a popular benchmark dataset among traditional hand segmentation works, it serves the purpose for comparing our deep learning segmentation approach against them.}


\subsection{{Experimental setup}}\label{Subsec:experimental-Setting}
The Keras open source deep learning library \cite{Keras} with a TensorFlow \cite{tensorfelow} backend was used to implement the proposed method. {As explained} in Section \ref{Subsec:training-Paradigm}, we {trained} our HGR-Net in three main steps. For each step, we {adopted} the same hyperparameters, excluding batch size {and dropout rates}. We used cross-entropy loss (Eq. \ref{Eq:: CE-loss}) as the cost functions, {and applied} the {Adam} optimization algorithm \cite{kingma2014} to minimize it. The learning rate is initialized as 0.001, and $\beta_{1}=0.9$ and $\beta_{2}=0.999$.
We set {the mini-batch size} as 8 for the first step (training the segmentation network) and 2 for {the other} steps. {The dropout rates for each training step are summarized in Table \ref{Table:drop}.}

\begin{table}[!t]
\caption{{\scriptsize\textit{Dropout rates for each training step of HGR-Net.}} \label{Table:drop}}
\scalebox{0.76}
{\begin{tabular*}{25pc}{@{\extracolsep{\fill}}lllc@{}}\toprule
{\bf {Layer}}   & {\bf {Training}} &  {\bf {Dropout Rate Description}}   &  {\bf {Rate}} \\ 
 \midrule
{dropout1}    & {Step 1}  & {before Conv 1$\times$1} & {0.20}
\\
 \midrule


{dropout2}    &  {Step 2}  & {before fc1 layer (both stream networks)}  & {0.20}  \\
{dropout3}    & {Step 2}  & {after fc1 layer (both stream networks)}  & {0.30}  \\
\midrule

{dropout4}    &  {Step 3}  & {after fc2 layer (appearance stream)}   & {0.75}  \\
{dropout5}    &  {Step 3}  & {after fc2 layer (shape stream)}   & {0.45}  \\
{dropout6}    &  {Step 3}  & {after fusion layer}  & {0.45}  \\
\midrule

\end{tabular*}}{}
\end{table}
The proposed framework {was} trained using an Nvidia GTX 1080TI GPU with 3.00GHz 4-core CPU and 16GB RAM, {under Cuda 9.0 with cuDNN 7.0.5 on TensorFlow GPU 1.5.0}. The maximum
number of epochs for each training step {was} fixed at 150, and the model with the
best validation accuracy was selected and {used for evaluation}. The training time was {around 12 hours for all steps}. Our trained HGR-Net takes roughly 23ms for a $320\times320$ input {RGB frame}. This shows that our model {has the potential} for real-time implementation.

  \begin{figure}[h]
\centering
\includegraphics[scale=0.28]{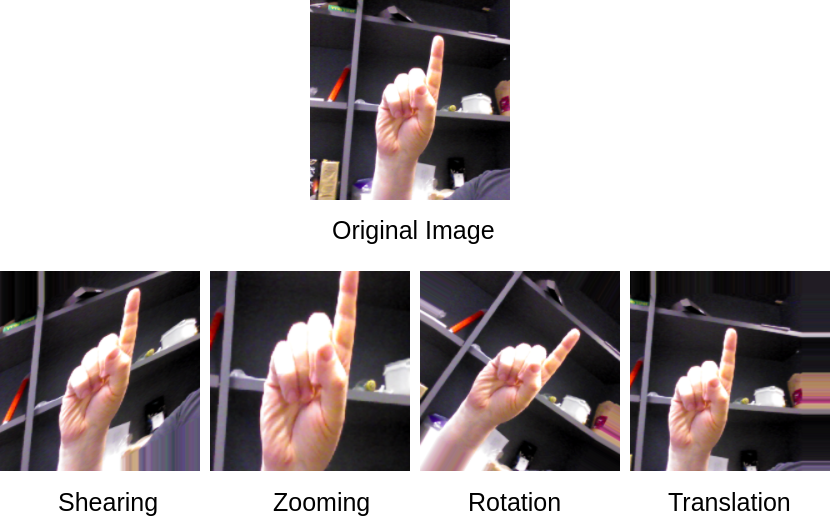}
\caption{\scriptsize\textit{{The transformations of the original data used in online data augmentation.}} } \label{fig:DA}
\end{figure}

 \subsection{{Data Augmentation}}\label{Subsec:dataAugmentation}

To address the problem of dataset bias, we apply both online and offline data augmentation to prevent overfitting and to improve the model generalization.
Our offline data augmentation consists of two transformations: zooming (15\%-20\%) and vertical/horizontal translations (15\%-20\%). With these operations, we increase the training set 4-fold.

\begin{table}[!b]
\caption{\scriptsize\textit{Transformation values for online data augmentation {during the training steps}}.\label{Table:data-augmentation}}
\scalebox{0.8}{
{\begin{tabular*}{22pc}{@{\extracolsep{\fill}}lllll@{}}\toprule
 &{ \bf rotation}  & {\bf shearing} & {\bf zooming} & {\bf translation} \\

\midrule
stream networks & $\pm$ 36$^\circ$    & 20\%     & $\pm$ $10$\%    & 2\% \\

{HGR-Net}  & $\pm$ 41$^\circ$     & 25\%     & $\pm$ $15$\%    & 15\%    \\ 
\bottomrule
\end{tabular*}}{}}
\end{table}

In online data augmentation, we augment data during training by applying 
rotation, shearing, zooming, and vertical/horizontal translations (Fig. \ref{fig:DA}).
All of these operations are applied randomly inside the mini-batches that are fed into the model. The specified value for each transformation is shown in Table \ref{Table:data-augmentation}. Since we train our {shape and appearance} stream networks from scratch, we adopt lower transformation values for these steps which helps the models to converge faster.

\subsection{{Experiments on Hand Segmentation}}\label{Subsec:HanSeg_experimental}
 We evaluated our hand segmentation model on OUHANDS \cite{Matilainen2016} and HGR1 \cite{HGR1} datasets. To carry out a quantitative comparison of accuracy of the proposed method, we measure {\it F-score} which is the equally weighted average of the precision and recall.

First, we evaluate the performance of {the segmentation stage of HGR-Net (Stage 1)} with and without the ASPP module. {{We also} employ the original architectures of FCN-8s \cite{Long2015}, PSPNet \cite{Zhao2017PyramidSP}, and DeepLabv3 \cite{Chen2017}, to compare their performance with our segmentation model on this specific task. The results are reported in Table \ref{Table:Seg-eval}. We observe that our model works better when used in conjunction with the ASPP module, which is approximately as accurate as DeepLab3 and PSPNet, while being much smaller and 2X faster {at run-time} for a $320 \times 320$ RGB image.
We use ResNet-50 \cite{He2016} as feature extractor for both DeepLabv3 and PSPNet architectures. Our model also slightly outperforms FCN-8s, while {also} being much smaller and 3X faster.} {Our proposed model requires the least amount of training time in comparison to all the other networks in Table \ref{Table:Seg-eval} given the fraction of parameters that need to be trained.} 

\begin{figure}[t]
\centering
\includegraphics[scale=0.18]{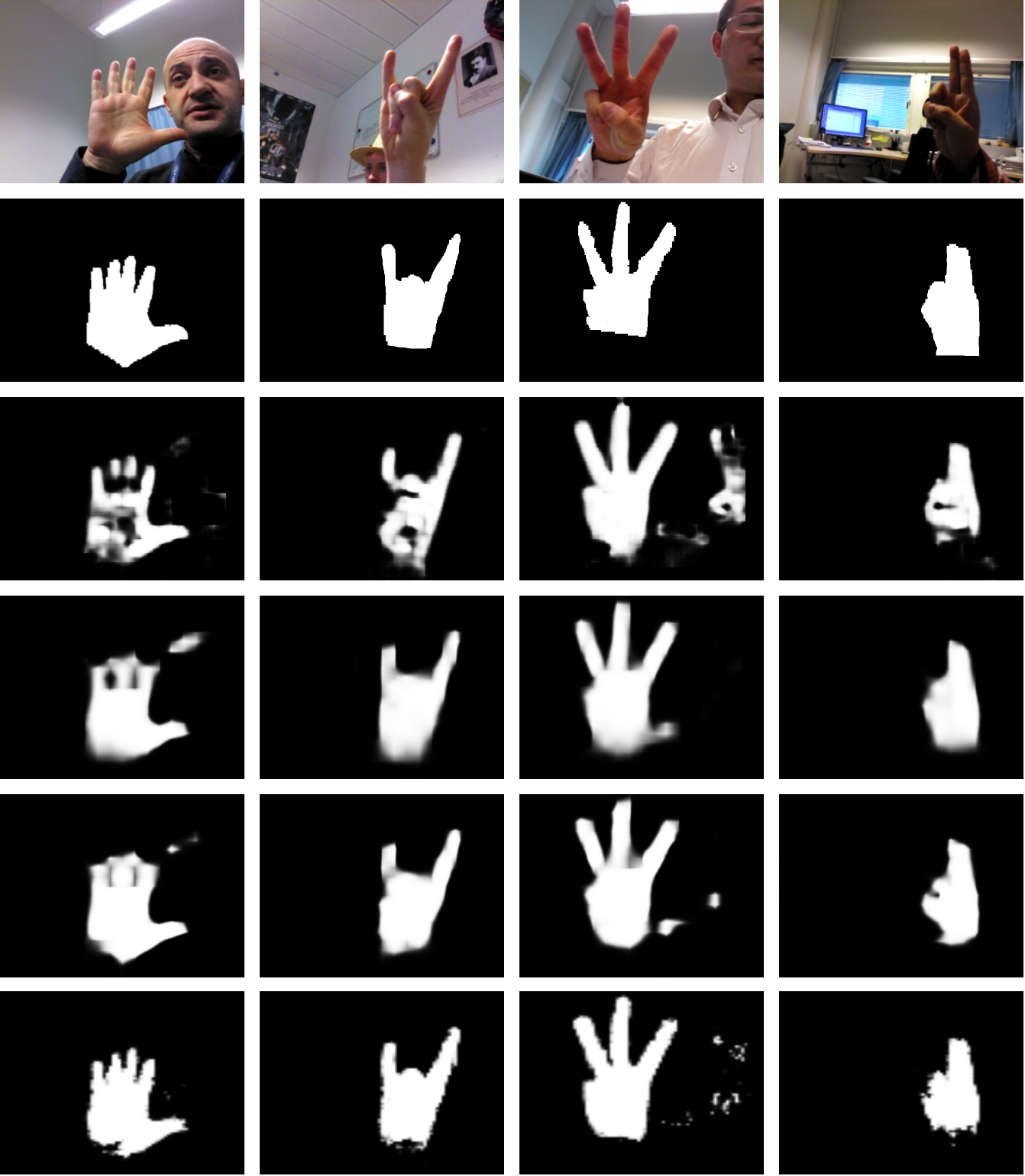}
\caption{\scriptsize\textit{{Example segmentations on OUHANDS test set: (top to bottom rows) Input images, ground truth, FCN-8s \cite{Long2015}, PSPNet \cite{Zhao2017PyramidSP}, DeepLabv3 \cite{Chen2017}, and HGR-Net (Stage 1).}}}  \label{fig:SegResult-ouhands}
\end{figure}

 \begin{table}[!b]
\caption{\scriptsize\textit{{Segmentation results on OUHANDS test set.\label{Table:Seg-eval}}}}
\scalebox{0.7}{
{\begin{tabular*}{25pc}{@{\extracolsep{\fill}}lllll@{}}\toprule
{\bf Method}  &{\it F-score}  & {\bf Time} & {\bf \#Parameters} & {\bf Model}  \\
  &  &  {\bf (ms)} & & {\bf Size} \\
  \midrule
        FCN-8s \cite{Long2015} & {0.9559}&63             & 134M & 537MB \\  
        PSPNet \cite{Zhao2017PyramidSP}& 0.9702 &50      &  79.44M &318MB\\
        DeepLabv3 \cite{Chen2017}& {\bf 0.9735} & 43     &  75.30M &302MB\\
        HGR-Net {\tiny (Stage 1-no ASPP)} & 0.9529& {\bf 20} &  {\bf 0.13M} & {\bf 0.79MB}\\  
       HGR-Net {\tiny (Stage1)} & {0.9630}& 21           & 0.28M  &1.4MB\\  
\bottomrule
\end{tabular*}}{}}
\end{table}

 
{To demonstrate the efficacy of our model qualitatively, some comparative segmentation results on {the OUHANDS test set} are illustrated in Fig. \ref{fig:SegResult-ouhands}.} {The images show} that our segmentation model deals well with common challenges in hand segmentation, such as uncontrolled backgrounds and lighting conditions. 

We further use the HGR1 dataset to compare our hand segmentation model against state-of-the-art methods which are based on traditional feature extraction. To evaluate our model on HGR1, we use 3-fold cross-validation. 
In each fold, we randomly select 66\% of the images for training and the rest for validation. We also perform offline data augmentation (Section \ref{Subsec:dataAugmentation}) on each training set and increase them 4-fold.
{As shown in Table \ref{Table: Seg-eval-hgr1},  HGR-Net (stage 1)  outperforms these traditional methods while it obtains a very comparable performance against DeepLabv3 and PSPNet at a fraction of the {run}-time, model size, and number of parameters required}.

\begin{figure*}[!t]
\centering
\includegraphics[scale=0.25]{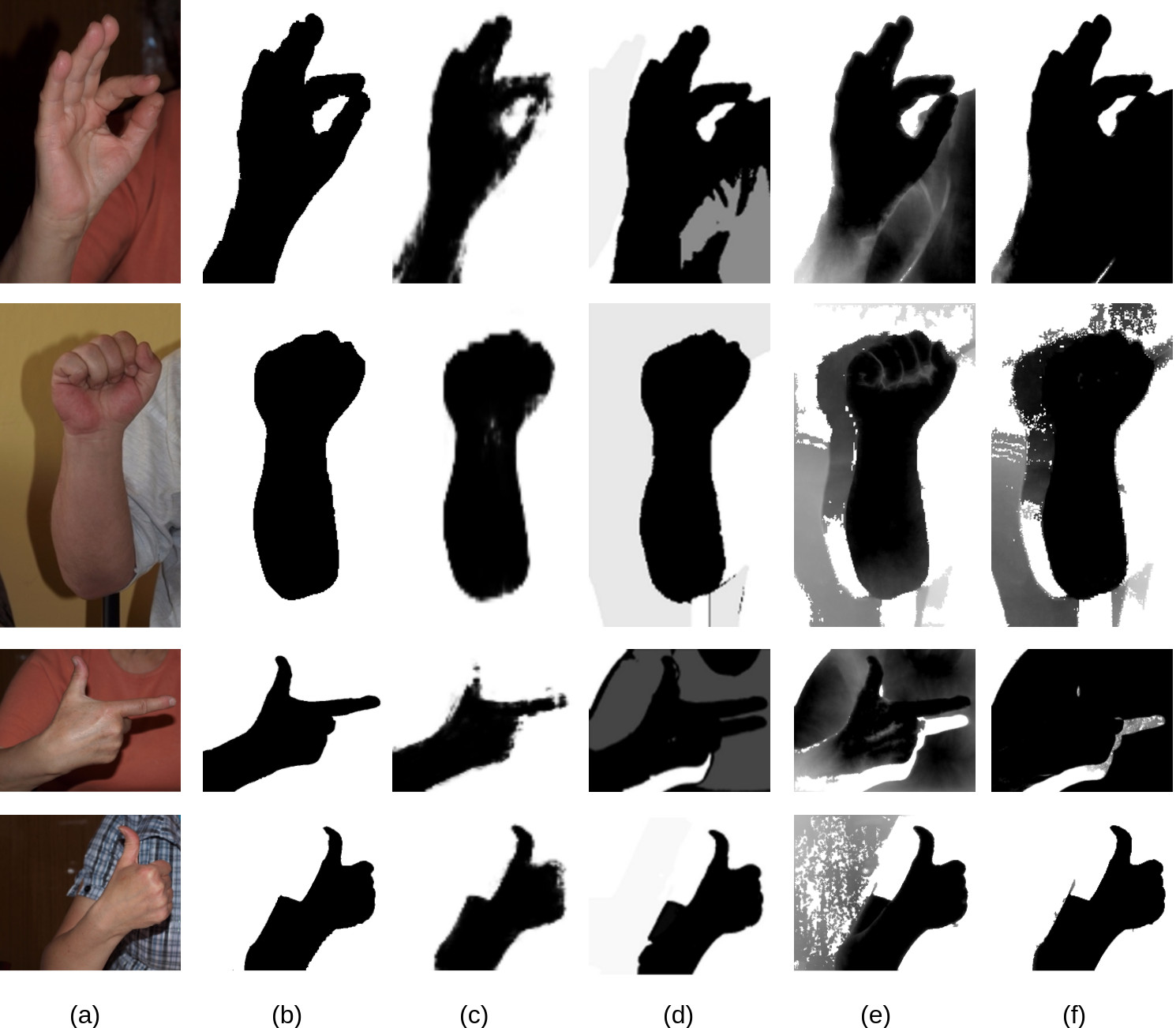}
\caption{\scriptsize\textit{{Comparison of segmentation results on HGR1 dataset. (a) Input image, (b) ground truth, (c) HGR-Net (Stage 1). (d) Hettiarachchi \textit{et al} \cite{Hettiarachchi2016}, (e) Kawulok \textit{et al}, \cite{Kawulok2014}, (f) Kawulok \cite{Kawulok2013}. Note, these images were selected from the validation set.}}} \label{fig:SegResults-HGR1}
\end{figure*}

  \begin{table}[!b]
\caption{\scriptsize\textit{{Segmentation results on HGR1 dataset.\label{Table: Seg-eval-hgr1}}}}
\scalebox{0.7}{
{\begin{tabular*}{25pc}{@{\extracolsep{\fill}}lllll@{}}\toprule
{\bf Method}  &{\it F-score}  & {\bf Time} & {\bf \#Parameters} & {\bf Model}  \\
  &  &  {\bf (ms)} & & {\bf Size} \\
\midrule
        Hettiarachchi \textit{et al.} \cite{Hettiarachchi2016} & 0.9692 &-& -&- \\
        Kawulok \textit{et al}. \cite{Kawulok2014}  & 0.9562 &- & - &- \\
      
        Kawulok \cite{Kawulok2013} & 0.9086 &-& -&- \\
        FCN-8s \cite{Long2015}  &   0.9772  & 63  & 134M &537MB \\
        PSPNet \cite{Zhao2017PyramidSP}  &  0.9877  & 50 &79.44M &318MB \\
        DeepLabv3 \cite{Chen2017} & {\bf 0.9883} & 43   &75.30M & 302MB\\
       {HGR-Net (Stage 1)} & {0.9825} &  {\bf 21}    & {\bf 0.28M} & {\bf 1.4MB} \\
\bottomrule
\end{tabular*}}{}}
\end{table}

From the qualitative results shown in Fig. \ref{fig:SegResults-HGR1}, we can see that traditional methods tend to fail in distinguishing skin pixels from skin-like backgrounds due to {the limited ability of their}  low-level, hand-crafted features.

%
\subsection{{Experiments on HGR}} \label{Subsec:HanRec_experimental}
{Next, we } evaluate HGR-Net for the task of hand gesture recognition. Table \ref{Table:exp-dataAugmentation} shows the classification scores on {the OUHANDS test set} with different forms of data augmentation.
{As evident} in Table \ref{Table:exp-dataAugmentation}, online augmentation provides higher test accuracy than offline, but the highest test accuracy is achieved when we trained HGR-Net using both offline and online. These results confirm that this data augmentation strategy can significantly reduce overfitting and improves the generalization capability of the model. Hence, it can be useful for robust hand gesture recognition in real-world scenes.

\begin{table}[!b]
\caption{\scriptsize\textit{Evaluation of HGR-Net for hand gesture recognition task with different forms of data augmentation}\label{Table:exp-dataAugmentation}}
\scalebox{0.7}{
{\begin{tabular*}{25pc}{@{\extracolsep{\fill}}lll@{}}\toprule
{\bf Data Augmentation}  &{\bf Train}  & {\bf Test} {({\it F-score})}
 \\
\midrule
        None & 0.9787 & 0.7842 \\
        Offline & 0.9760 & 0.8165 \\
        Online & 0.9000 & 0.8629 \\
        Offline+Online &0.8742& \bf{0.8810}\\
\bottomrule
\end{tabular*}}{}}
\end{table}

\begin{table*}[!h]
\caption{\scriptsize\textit{Comparison of recognition accuracy on OUHANDS test set. { Note that for fair comparison we use the proposed data augmentation strategy for training all models considered in the table}}\label{Table:exp-recognition}}
\scalebox{1}{
{\begin{tabular*}{\textwidth}{@{\extracolsep{\fill}}llccccc}\toprule
\# & \bf{ Model}     & \bf{{\it F-score} } & \bf{Input Size} &\bf{Time (ms)} &\bf{\#Parameters}& \bf{Model Size} \\
\midrule
 \multicolumn{2}{c}{ Input: RGB}  \\
        \midrule

1& HGR-Net {\tiny (Shape stream only)}   &  0.8527  & 320$\times$320   & 22 & 0.385M & 1.9MB \\ 

2& HGR-Net {\tiny (Appearance stream only)}     &  0.7545  &  320$\times$320  & 0.7 &  0.106M & 0.45MB\\       
3& HGR-Net     &  \bf{0.8810}  &  \bf{320$\times$320}  &  \bf{23} & \bf{0.499M} & \bf{2.4MB} \\ 
\midrule

       4& ResNet-50 \cite{He2016}                 &  0.8138  &  224$\times$224 & 25 &23.60M& 99MB  \\ 
       5& DenseNet-121 \cite{Huang2017DenselyCC} &0.8281    &  224$\times$224 & 24 &7.04M & 33MB \\ 
       6& MobileNet \cite{howard2017mobilenets}  &0.8650    &  224$\times$224 & 13 &3.22M& 16MB \\

\midrule

  \multicolumn{2}{c}{ Input: binary}  \\

     \midrule
       7& CNN1 \cite{Oyedotun2017}  &0.8667 &  32$\times$32  & 0.3 & 0.126M &0.51MB \\ 
       8& {HGR-Net {\tiny (Shape stream, no Stage 1)}} &  \bf{0.9375 } &  \bf{320$\times$320}  & \bf{0.4} & \bf{0.106M} & \bf{0.44MB}\\
\bottomrule
\end{tabular*}}{}}
\end{table*}

Table \ref{Table:exp-recognition} compares the performances of different network architectures on the OUHANDS dataset. 
{Rows 1-3 exhibit the influence of each network stream in HGR-Net's performance.}
The performance of HGR-Net as a whole then jumps to an {\it F-score} of $0.8810$ when we
apply our late fusion approach to coalesce the shape and appearance streams. This result emphasizes the effectiveness of our fusion network architecture.

{In rows 4-6 of Table \ref{Table:exp-recognition}, we present the scores} of three popular deep architectures, ResNet-50 \cite{He2016}, DenseNet-121 \cite{Huang2017DenselyCC} and MobileNet \cite{howard2017mobilenets}, for comparison. {For training these networks on the OUHANDS dataset, we only replace their softmax layer, originally trained to recognize 1000 classes, with a softmax layer that recognizes only 10 (i.e. the ten classes in OUHANDS). The weights of other layers are {initialized} from the original network that was trained on ImageNet \cite{deng2009imagenet}.} {Note that the networks' layers are not frozen during the training process.} {It can be seen} that our full HGR-net (row 3) outperforms these models, {and obtains a 1.6\% increment over the next best, i.e. MobileNet, while being 6X smaller {and bearing a fraction of its parameters}. It is interesting to note that we do not use any large-scale dataset for pre-training our model. This can be useful in real-world applications such as online learning in which the HGR system relies on a limited amount of in-coming training data. Moreover, our lightweight CNN is more feasible to deploy on hardware with limited memory.} {The proposed approach operates at approximately 43 fps\footnote{{Real-time video demo: https://youtu.be/-KSNrk5plS0}}.}

{The last row of the table reports our model's performance on binary hand masks. 
Since we do not use RGB images in this experiment, we only employ the shape stream without relying on the segmentation maps {(i.e no part (a) in Figure \ref{fig:HGR-Net})}. 
As expected, our model then performs better than our model with RGB input (row 3) since it does not deal with image backgrounds in both training and {testing. For comparison, we} evaluate the performance of CNN1 \cite{Oyedotun2017} on OUHANDS. Since CNN1 is a single channel CNN, we train and test this network with only binary hand masks. As seen in {rows 7-8} of Table \ref{Table:exp-recognition}, our HGR-Net with binary input outperforms CNN1 by a significant margin of $\sim$6\%).} 

\begin{figure}[h]
\centering
\includegraphics[scale=0.38]{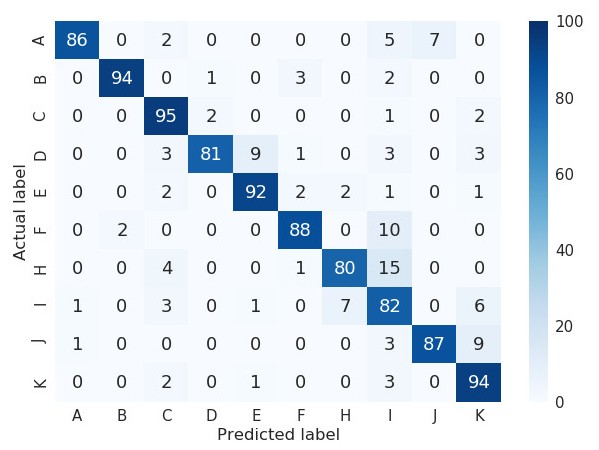}
\caption{\scriptsize\textit{{Confusion matrix of the proposed method for hand gesture recognition task on the OUHANDS test set. }}}  \label{fig:confusion-matrix}
\end{figure}

Fig. \ref{fig:confusion-matrix} shows the confusion matrix of the full HGR-Net in terms of recognition accuracy on OUHANDS test set. The {column and row indices} denote the predicted results and the target classes, respectively.

\section{{Conclusion}}\label{Sec:Conclusion}
{We presented} a two-stage fusion network, based on CNN architecture for the problem of hand gesture recognition. To improve the recognition performance without using depth data, we proposed a hand segmentation architecture in the first stage of the network. Our experimental results showed that our hand segmentation model has a {great} performance against challenging situations, even when the background color is similar to skin color. For the second stage of the network, we designed a two-stream CNN which can learn to fuse feature representations from both {the RGB image and its} segmentation map before classification. Moreover, we employed an effective data augmentation technique which plays an important role in obtaining higher recognition accuracy. Our best proposed model
achieves state-of-the-art performance on the OUHANDS dataset {with a real-time average recognition speed of 23ms per frame or 43 fps}.
In addition, we evaluated our segmentation model on the HGR1 dataset {where} it outperformed traditional state-of-the-art methods.


\end{document}